\pdfoutput=1 

\documentclass[10pt,twocolumn,letterpaper]{article}

\usepackage[table]{xcolor}

\usepackage{tikz}
\usepackage{cvpr}
\usepackage{times}
\usepackage{epsfig}
\usepackage{graphicx}
\usepackage{amsmath}
\usepackage{amssymb}

\usepackage{subcaption}
\usepackage{soul}
\usepackage{booktabs}
\usepackage[utf8]{inputenc}

\newif\ifdraft
\draftfalse
\drafttrue

\definecolor{orange}{rgb}{1,0.5,0}
\definecolor{violet}{RGB}{70,0,170}

\ifdraft
 \newcommand{\PF}[1]{{\color{red}{\bf PF: #1}}}
 
 \newcommand{\KY}[1]{{\color{blue}{\bf KY: #1}}}
 
 \newcommand{\MS}[1]{{\color{green}{\bf MS: #1}}}
 
 \newcommand{\ZD}[1]{{\color{violet}{\bf ZD: #1}}}
 
 \newcommand{\YH}[1]{{\color{orange}{\bf YH: #1}}}
 
\else
 \newcommand{\PF}[1]{}
 
 \newcommand{\KY}[1]{}
 
 \newcommand{\MS}[1]{}
 
 \newcommand{\ZD}[1]{}
 
 \newcommand{\YH}[1]{}
 
\fi

\newcommand{\comment}[1]{}
\newcommand{\parag}[1]{\vspace{-3mm}\paragraph{#1}}

\usepackage[pagebackref=true,breaklinks=true,letterpaper=true,colorlinks,bookmarks=false]{hyperref}

\cvprfinalcopy % *** Uncomment this line for the final submission

\def\cvprPaperID{1717} % *** Enter the CVPR Paper ID here

\ifcvprfinal\fi
\begin{document}

%%%%%%%%% TITLE
\title{Segmentation-driven 6D Object Pose Estimation}

\author{
  \vspace{0.5em}
  {Yinlin Hu, \quad Joachim Hugonot, \quad Pascal Fua, \quad Mathieu Salzmann} \\
  {CVLab, EPFL, Switzerland} \\
  {\tt\small \{yinlin.hu, joachim.hugonot, pascal.fua, mathieu.salzmann\}@epfl.ch} \\
}

\maketitle
\thispagestyle{empty}

% !TEX root = ../top.tex
% !TEX spellcheck = en-US

%%%%%%%%% ABSTRACT
\begin{abstract}

The most recent trend in estimating the 6D pose of rigid objects has been to train deep networks to either directly regress the pose from the image or to predict the 2D locations of 3D keypoints, from which the pose can be obtained using a PnP algorithm. In both cases, the object is treated as a global entity, and a single pose estimate is computed. As a consequence, the resulting techniques can be vulnerable to large occlusions. 

In this paper, we introduce a segmentation-driven 6D pose estimation framework where each visible part of the objects contributes a local pose prediction in the form of 2D keypoint locations. We then use a predicted measure of confidence to combine these pose candidates into a robust set of 3D-to-2D correspondences, from which a reliable pose estimate can be obtained. We outperform the state-of-the-art on the challenging Occluded-LINEMOD and YCB-Video datasets, which is evidence that our approach deals well with multiple poorly-textured objects occluding each other. Furthermore, it relies on a simple enough architecture to achieve real-time performance.

\end{abstract}
% !TEX root = ../top.tex
% !TEX spellcheck = en-US

\section{Introduction}

Image-based 6D object pose estimation is crucial in many real-world applications, such as augmented reality or robot manipulation. Traditionally, it has been handled by establishing correspondences between the object's known 3D model and 2D pixel locations, followed by using the Perspective-n-Point (PnP) algorithm to compute the 6 pose parameters~\cite{Lepetit05b,Rothganger06,Wagner08}. While very robust when the object is well textured, this approach can fail when it is featureless or when the scene is cluttered with multiple objects occluding each other. 

\begin{figure}
\centering
\begin{tabular}{cc}
\includegraphics[width=0.45\linewidth,clip, trim=20 60 100 50]{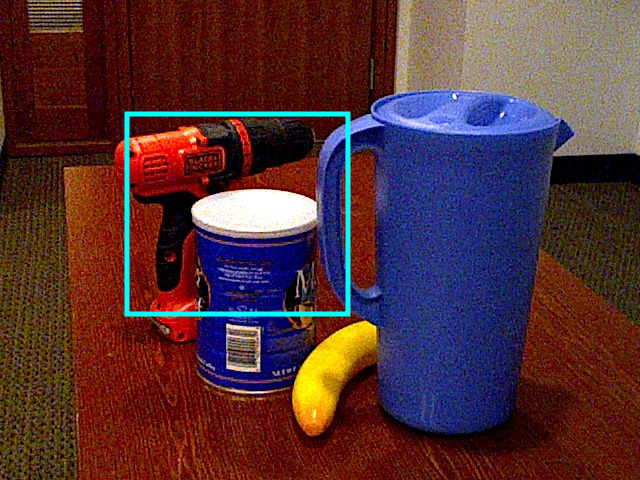}&
\includegraphics[width=0.45\linewidth,clip, trim=20 60 100 50]{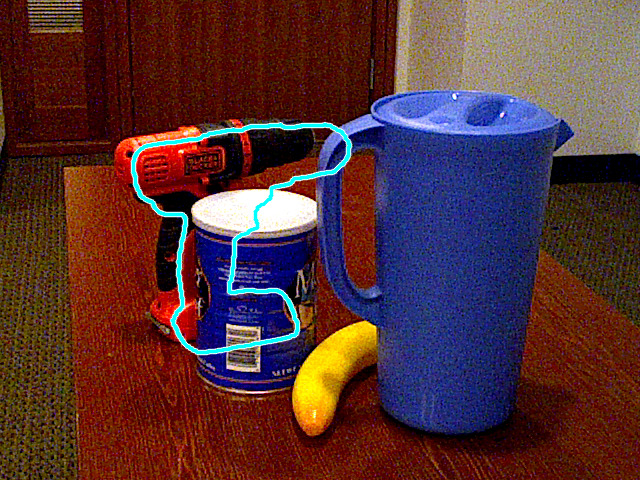}\\
(a)&(b)\\
\includegraphics[width=0.45\linewidth,clip, trim=20 60 100 50]{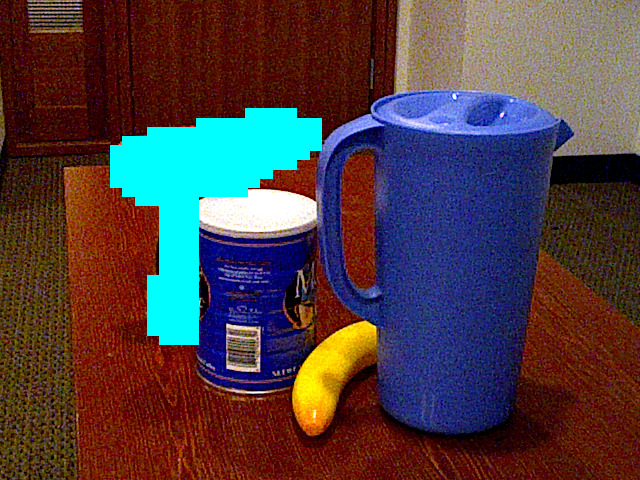}&
\includegraphics[width=0.45\linewidth,clip, trim=20 60 100 50]{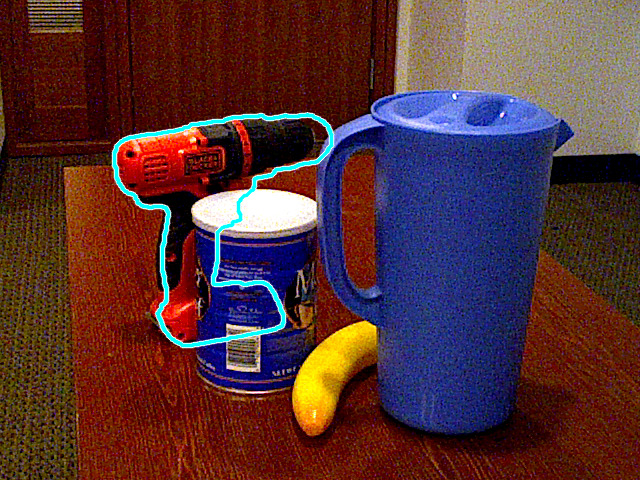}\\
(c)&(d)
\end{tabular}
\vspace{-3mm}
\caption{ \small {\bf Global pose estimation vs our segmentation-driven approach.}  {\bf (a)} The drill's bounding box overlaps another occluding it. {\bf (b)} As a result, the globally-estimated pose~\cite{Xiang18b} is wrong. {\bf (c)} In our approach, only image patches labeled as corresponding to the drill contribute to the pose estimate. {\bf (d)} It is now correct.}
\label{fig:teaser}
\end{figure}

Recent work has therefore focused on overcoming these difficulties, typically using deep networks to either regress directly from image to 6D pose~\cite{Kehl17,Xiang18b} or to detect keypoints associated to the object~\cite{Rad17,Tekin18a}, which can then be used to perform PnP.  In both cases, however, the object is still treated as a global entity, which makes the algorithm vulnerable to large occlusions. Fig.~\ref{fig:teaser} depicts a such a case: The bounding box of an occluded drill overlaps other objects that provide irrelevant information to the pose estimator and thereby degrade its performance. Because this happens often, many of these recent methods require an additional post-processing step to refine the pose~\cite{Li18a}.

% !TEX root = ../top.tex
% !TEX spellcheck = en-US

In this paper, we show that more robust pose estimates can be obtained by combining multiple local predictions instead of a single global one.
To this end, we introduce a segmentation-driven 6D pose estimation network in which each visible object patch contributes a pose estimate for the object it belongs to in the form of the predicted 2D projections of predefined 3D keypoints. Using confidence values also predicted by our network, we then combine the most reliable 2D projections for each 3D keypoint, which yields a robust set of 3D-to-2D correspondences. We then use a RANSAC-based PnP strategy to infer a single reliable pose per object.

Reasoning in terms of local patches not only makes our approach robust to occlusions, but also yields a rough segmentation of each object in the scene. In other words, unlike other methods that divorce object detection from pose estimation~\cite{Rad17,Kehl17,Xiang18b}, we perform both jointly while still relying on a simple enough architecture for real-time performance.

In short, our contribution is a simple but effective segmentation-driven network that produces accurate 6D object pose estimates without the need for post-processing, even when there are multiple poorly-textured objects occluding each other. It combines segmentation and ensemble learning in an effective and efficient architecture. We will show that it outperforms the state-of-the-art methods on standard benchmarks, such as the OccludedLINEMOD and YCB-Video datasets.

% !TEX root = ../top.tex
% !TEX spellcheck = en-US

\begin{figure*}
	\centering
	\includegraphics[width=0.95\linewidth]{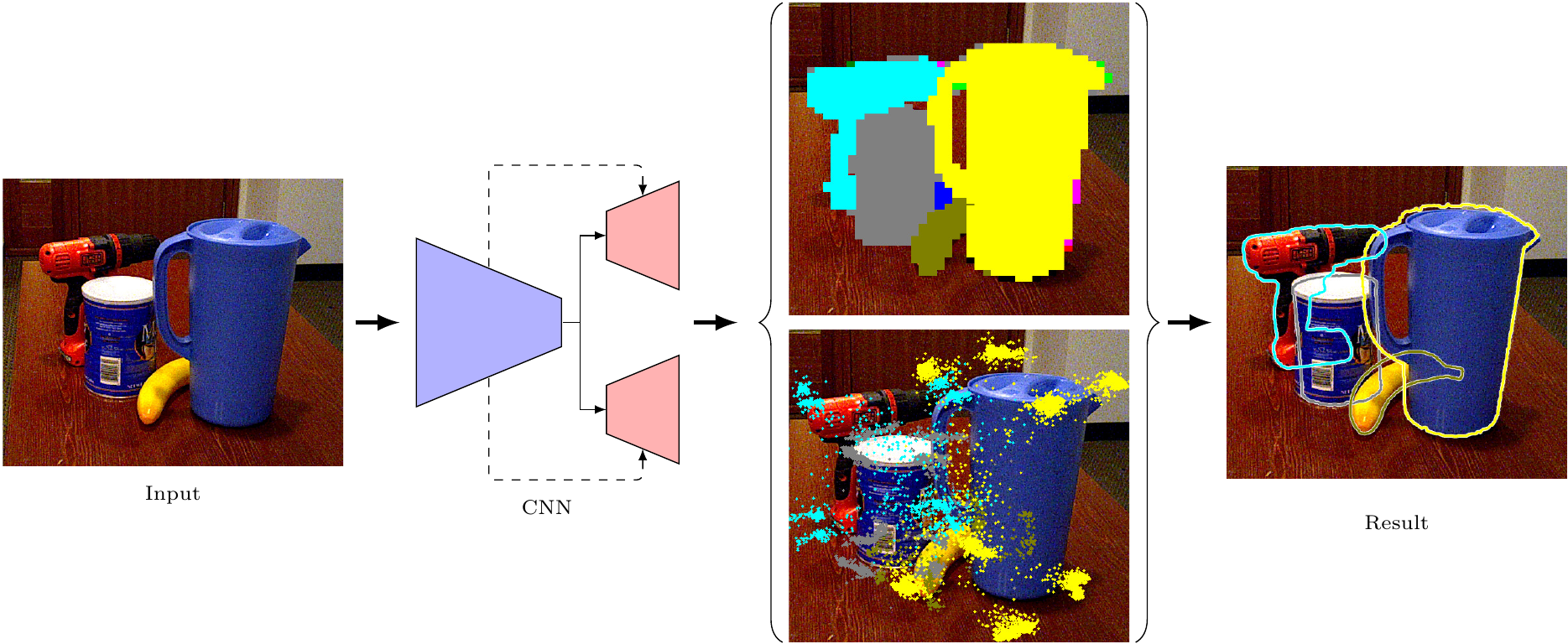}
	\vspace{-3mm}
	\caption{{\bf Overall workflow of our method.} Our architecture has two streams: One for object segmentation and the other to regress 2D keypoint locations. These two streams share a common encoder, but the decoders are separate. Each one produces a tensor of a spatial resolution that defines an $S\times S$ grid over the image. The segmentation stream predicts the label of the object observed at each grid location. The regression stream predicts the 2D keypoint locations for that object. 
}
	\label{fig:architecture}
\end{figure*}

\section{Related Work}
\label{sec:related}

In this paper, we focus on 6D object pose estimation from RGB images, without access to a depth map, unlike in RGBD-based methods~\cite{Hinterstoisser12b,Brachmann14,Brachmann16a,Michel17}. The classical approach to performing this task involves extracting local features from the input image, matching them with those of the model, and then running a PnP algorithm on the resulting 3D-to-2D correspondences. Over the years, much effort has been invested in designing local feature descriptors that are invariant to various transformations~\cite{Lowe04,Tola10,Trzcinski12c,Tulsiani15,Pavlakos17a,Ono18}, so that they can be matched more robustly~\cite{Muja09,Muja14,Hu16}. In parallel, increasingly effective PnP methods have been developed to handle noise and mismatches~\cite{Lepetit09,Zheng13,Li12c,Ferraz14}. As a consequence, when dealing with well-textured objects, feature-based pose estimation is now fast and robust, even in the presence of mild occlusions. However, it typically struggles with heavily-occluded and poorly-textured objects. 

In the past, textureless objects have often been handled by template-matching~\cite{Hinterstoisser12a,Hinterstoisser12b}. Image edges then become the dominant information source~\cite{li11,Lowe91}, and researchers have developed strategies based on different distances, such as the Hausdorff~\cite{Huttenlocher93} and the Chamfer~\cite{liu10a,Hsiao14a} ones, to match the 3D model against the input image. While effective for poorly-textured objects, these techniques often fail in the presence of mild occlusions and cluttered background.

As in many computer vision areas, the modern take on 6D object pose estimation involves deep neural networks. Two main trends have emerged: Either regressing from the image directly to the 6D pose~\cite{Kehl17,Xiang18b} or predicting 2D keypoint locations in the image~\cite{Rad17,Tekin18a}, from which the pose can be obtained via PnP. Both approaches treat the object as a global entity and produce a single pose estimate.
% for the whole of it.  
This makes them vulnerable to occlusions because, when considering object bounding boxes as they all do, signal coming from other objects or from the background will contaminate the prediction. While, in~\cite{Rad17,Xiang18b}, this is addressed by segmenting the object of interest, the resulting algorithms still provide a single, global pose estimate, that can be unreliable, as illustrated in Fig.~\ref{fig:teaser} and demonstrated in the results section.  As a consequence, these methods typically invoke an additional pose refinement step~\cite{Li18a}. 
%\PF{Is this shown in the teaser figure as well?}\YH{I skip adding this to the teaser figure, maybe making the teaser figure focused on the comparison of bounding box and segmentation mask can make our title xxx-driven more soundable.}

To the best of our knowledge, the work of~\cite{Jafari18} and~\cite{Oberweger18} constitute the only recent attempts at going beyond a global prediction. While the method in~\cite{Jafari18} also relies on segmentation via a state-of-the-art semantic segmentation network, its use of regression to 3D object coordinates, which reside in a very large space, yields disappointing performance. 
%the semantic information for 6D pose estimation. However, although they use a state-of-the-art semantic segmentation method as their first stage, the multi-stage structure and the regression target of enormous 3D space make their performance rather worse than expected. 
By contrast, the technique in~\cite{Oberweger18} predicts multiple keypoint location heatmaps from local patches and assembles them to form an input to a PnP algorithm. The employed patches, however, remain relatively large, thus still potentially containing irrelevant information. Furthermore, at runtime, this approach relies on a computationally-expensive sliding-window strategy that is ill-adapted to real-time processing.
Here, we propose to achieve robustness by combining multiple local pose predictions in an ensemble manner and in real time, without post-processing. In the results section, we will show that this outperforms the state-of-the-art approaches~\cite{Kehl17,Xiang18b,Rad17,Tekin18a,Jafari18,Oberweger18}. 

Note that human pose estimation~\cite{He17,Wei16,Papandreou18} is also related to global 6D object pose prediction techniques.
By targeting non-rigid objects, however, these methods require the more global information extracted from larger receptive fields and are inevitably more sensitive to occlusions.
By contrast, dealing with rigid objects allows us to rely on local predictions that can be robustly combined, and local visible object parts can provide reliable predictions for all keypoints. We show that assembling these local predictions yields robust pose estimates, even when observing multiple objects that occlude each other.

\section{Approach}
\label{sec:approach}

Given an input RGB image, our goal is to simultaneously detect objects and estimate their 6D pose, in terms of 3 rotations and 3 translations. We assume the objects to be rigid and their 3D model to be available. As in~\cite{Rad17,Tekin18a}, we design a CNN architecture to regress the 2D projections of some predefined 3D points, such as the 8 corners of the objects' bounding boxes. However, unlike these methods whose predictions are global for each object and therefore affected by occlusions, we make individual image patches predict both to which object they belong and where the 2D projections are. We then combine the predictions of all patches assigned to the same object for robust PnP-based pose estimation. 

Fig.~\ref{fig:architecture} depicts the corresponding workflow.  In the remainder of this section, we first introduce our two-stream network architecture. We then describe each stream individually and finally our inference strategy.

\subsection{Network Architecture}
\label{eq:network}

In essence, we aim to jointly perform segmentation by assigning image patches to objects and  2D coordinate regression of keypoints belonging to these objects, as shown in Fig.~\ref{fig:position_prediction}. To this end,  we design the two-stream architecture depicted by Fig.~\ref{fig:architecture}, with one stream for each task. It has an encoder-decoder structure, with a common encoder for both streams and two separate decoders. 

For the encoder, we use the  Darknet-53 architecture of YOLOv3~\cite{Redmon18} that has proven highly effective and efficient for objection detection. For the decoders, we designed networks that output 3D tensors of spatial resolution $S\times S$ and feature dimensions $D_{seg}$ and $D_{reg}$, respectively. This amounts to superposing an $S\times S$ grid on the image and computing a feature vector of dimension $D_{seg}$ or $D_{reg}$ per grid element. The spatial resolution of that grid controls the size of the image patches that vote for the object label and specific keypoint projections. A high resolution yields fine segmentation masks and many votes. However, it comes at a higher computational cost, which may be unnecessary for our purposes. Therefore, instead of matching the 5 downsampling layers of the Darknet-53 encoder with 5 upsampling layers, we only use 2 such layers, with a standard stride of 2. The same architecture, albeit with a different output feature size, is used for both decoder streams.

To train our model end-to-end, we define a loss function
\begin{equation}
{\mathcal L}={\mathcal L}_{seg} + {\mathcal L}_{reg}\;,
\label{eq:lossF}
\end{equation}
which combines a segmentation and a regression term that we use to score the output of each stream. We now turn to their individual descriptions.  

\subsection{Segmentation Stream}
\label{sec:seg}

The role of the segmentation stream is to assign a label to each cell of the virtual $S \times S$ grid superposed on the image, as shown in Fig.~\ref{fig:position_prediction}(a). 
%using fully-convolutional layers 
More precisely, given $K$ object classes, this translates into outputting a vector of dimension $D_{seg} = K+1$ at each spatial location, with an additional dimension to account for the background.

During training, we have access to both the 3D object models and their ground-truth pose. We can therefore generate the ground-truth semantic labels by projecting the 3D models in the images while taking into account the depth of each object to handle occlusions. In practice, the images typically contain many more background regions than object ones. Therefore, we take the loss ${\mathcal L}_{seg}$ of Eq.~\ref{eq:lossF} to be the Focal Loss of~\cite{Lin17}, a dynamically weighted version of the cross-entropy. Furthermore, we rely on the median frequency balancing technique of~\cite{Eigen15,Badrinarayanan15} to weigh the different samples. We do this according to the pixel-wise class frequencies rather than the global class frequencies to account for the fact that objects have different sizes.

\subsection{Regression Stream}
\label{sec:reg}

The purpose of the regression stream is to predict the 2D projections of predefined 3D keypoints associated to the 3D object models. Following standard practice~\cite{Rad17,Grabner18,Rad18}, we typically take these keypoints to be the 8 corners of the model bounding boxes. 

Recall that the output of the regression stream is a 3D tensor of size $S \times S \times D_{reg}$. Let $N$ be the number of 3D keypoints per object whose projection we want to predict. When using bounding box corners, $N=8$.
We take $D_{reg}$ to be $3N$ to represent at each spatial location the $N$ pairs of 2D projection values along with a confidence value for each. 

\input{fig/position_prediction}

In practice, we do not predict directly the keypoints' 2D coordinates. Instead, for each one, we predict an offset vector with respect to the center of the corresponding grid cell, as illustrated by Fig.~\ref{fig:position_prediction}(b). That is, let ${\bf c}$ be the 2D location of a grid cell center. For the $i^{th}$ keypoint, we seek to predict an offset ${\bf h}_i({\bf c})$, such that the resulting location ${\bf c} + {\bf h}_i({\bf c})$ is close to the ground-truth 2D location ${\bf g}_i$. During training, this is expressed by the residual 
\begin{equation}
{\Delta}_i({\bf c}) = {\bf c} + {\bf h}_i({\bf c}) - {\bf g}_i\;,
\end{equation}
and by defining the loss function
\begin{equation}
{\mathcal L}_{pos} = \sum_{{\bf c}\in M} \sum_{i=1}^{N} \Vert {\Delta}_i({\bf c}) \Vert_{1} \;,
\label{equ:xyregression}
\end{equation}
where $M$ is the foreground segmentation mask, and $\Vert\cdot\Vert_{1}$ denotes the $L^1$ loss function, which is less sensitive to outliers than the $L^2$ loss. Only accounting for the keypoints that fall within the segmentation mask $M$ focuses the computation on image regions that truly belong to objects.

As mentioned above, the regression stream also outputs a confidence value $s_i({\bf c})$ for each predicted keypoint, which is obtained via a sigmoid function on the network output. These confidence values should  reflect the proximity of the predicted 2D projections to the ground truth. To encourage this, we define a second loss term 
\begin{equation}
{\mathcal L}_{conf} = \sum_{{\bf c}\in M} \sum_{i=1}^{N} \big \Vert s_i({\bf c})  - \exp(-\tau \Vert {\Delta}_i({\bf c}) \Vert_{2} ) \big \Vert_{1}\;,
\label{equ:confregression}
\end{equation}
where $\tau$ is a modulating factor.  
%\PF{This sentence should be moved to where you describe the architecture. It does not belong here.}
We then take the regression loss term of Eq.~\ref{eq:lossF} to be
\begin{equation}
{\mathcal L}_{reg}= \beta{\mathcal L}_{pos} + \gamma{\mathcal L}_{conf}\;,
\label{equ:regloss}
\end{equation}
where $\beta$ and $\gamma$ modulate the influence of the two terms. 

Note that because the two terms in Eq.~\ref{equ:regloss} focus on the regions that are within the segmentation mask $M$, their gradients are also backpropagated to these regions only. As in the segmentation stream, to account for pixel-wise class imbalance, we weigh the regression loss term for different objects according to the pixel-wise class frequencies in the training set.

% !TEX root = ../top.tex
% !TEX spellcheck = en-US

\begin{figure}
	\centering
	\begin{tabular}{cc}
	\includegraphics[width=0.45\linewidth, clip, trim=80 20 140 100]{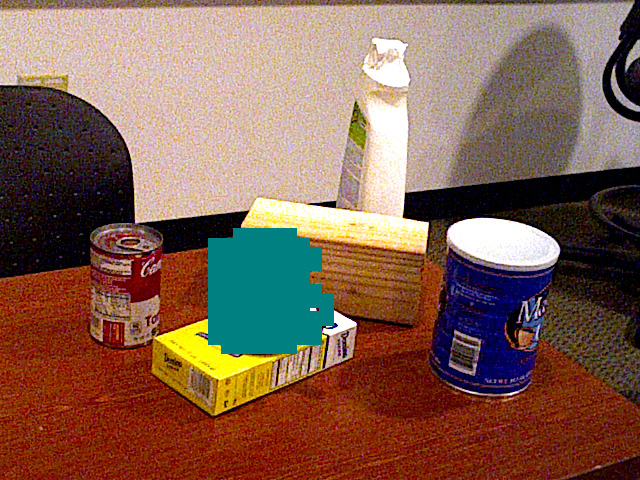}&
	\includegraphics[width=0.45\linewidth, clip, trim=80 20 140 100]{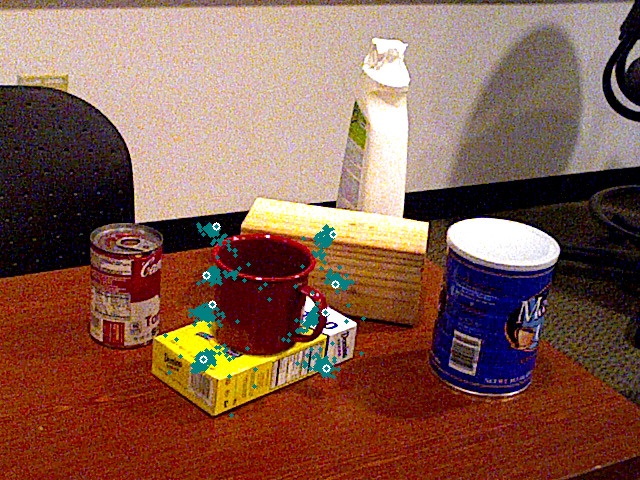}\\
	(a)&(b)\\
	\includegraphics[width=0.45\linewidth, clip, trim=80 20 140 100]{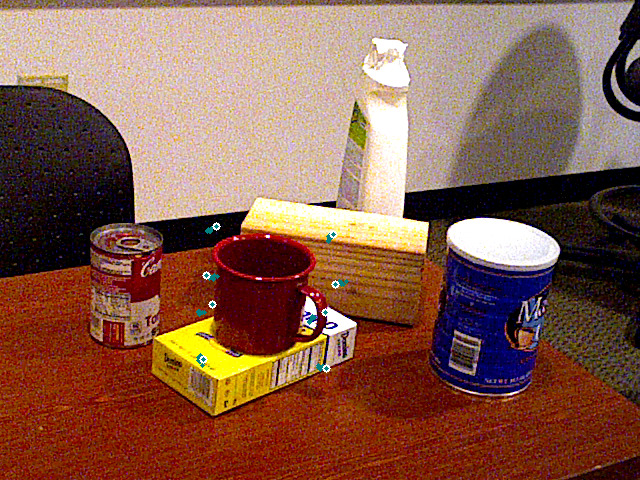}&
	\includegraphics[width=0.45\linewidth, clip, trim=80 20 140 100]{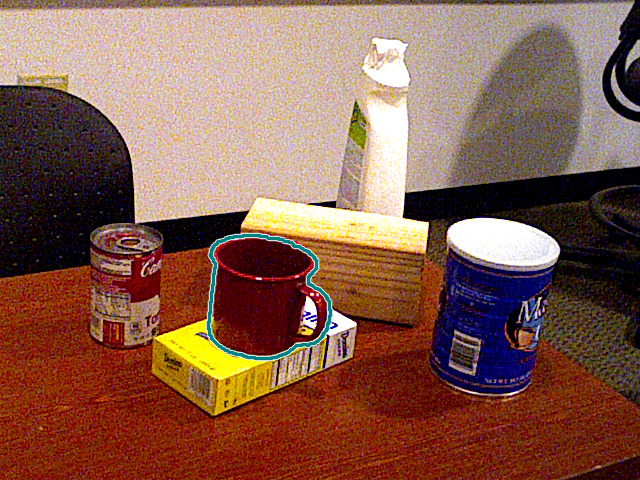}\\
	(c)&(d)
	\end{tabular}
	\vspace{-3mm}
	\caption{{\bf Combining pose candidates.} {\bf (a)} Grid cells predicted to belong to the cup are overlaid on the image. {\bf (b)} Each one predicts 2D locations for the corresponding keypoints, shown as green dots. {\bf (c)} For each 3D keypoint, the $n=10$ 2D locations about which the network is most confident are selected. {\bf (d)} Running a RANSAC-based PnP on these yields an accurate pose estimate, as evidenced by the correctly drawn outline. }	
		\label{fig:multi_candidates}
\end{figure}

% !TEX root = ../top.tex
% !TEX spellcheck = en-US

\begin{table*}
	\centering
	\rowcolors{3}{white}{gray!10}
	\begin{tabular}{lrrr!{\vrule width 1pt}rrrrrr}
		\toprule
		& \multicolumn{3}{c}{ADD-0.1d}	&	\multicolumn{6}{c}{REP-5px} \\
		&	PoseCNN	& Heatmaps & {\bf Ours} & 	PoseCNN	&	BB8	& Tekin & iPose & Heatmaps & {\bf Ours}	\\
		\midrule
		Ape			& 9.6  &{\bf 16.5}&12.1&34.6&28.5&7.0 & 24.2 &{\bf 64.7}&59.1\\
		Can			& {\bf 45.2}  &42.5&39.9&15.1&1.2&11.2 & 30.2 &53.0&{\bf 59.8}\\
		Cat 		& 0.9  &2.8&{\bf 8.2}&10.4&9.6&3.6& 12.3 & {\bf 47.9} & 46.9\\
		Driller		& 41.4  &{\bf 47.1}&45.2&7.4&0&1.4& - & 35.1&{\bf 59.0}\\
		Duck		& {\bf 19.6}  &11.0&17.2&31.8&6.8&5.1& 12.1 & 36.1&{\bf 42.6}\\
		Eggbox$^*$	& 22.0  &{\bf 24.7}&22.1&1.9& -  & -  & - & 10.3&{\bf 11.9}\\
		Glue$^*$	& 38.5  &{\bf 39.5}&35.8&13.8&4.7&6.5& 25.9 & {\bf 44.9}&16.5\\
		Holepun.	& 22.1  &21.9&{\bf 36.0}&23.1&2.4&8.3& 20.6 & 52.9&{\bf 63.6}\\
		\midrule
		Average		& 24.9	&25.8&{\bf 27.0}&17.2&7.6&6.2& 20.8 & 43.1&{\bf 44.9}\\
		\bottomrule
	\end{tabular}
	\vspace{-3mm}
	\caption{{\bf Comparison with the state of the art on Occluded-LINEMOD.} We compare our results with those of PoseCNN~\cite{Xiang18b}, BB8~\cite{Rad17}, Tekin~\cite{Tekin18a}, iPose~\cite{Jafari18}, and Heatmaps~\cite{Oberweger18}. The results missing from the original papers are denoted as ``-''.}
	\label{tab:occlinemod_eval}
\end{table*}

% !TEX root = ../top.tex
% !TEX spellcheck = en-US

\begin{table}
	\centering
\scalebox{0.85}{
	\begin{tabular}{lcccccc}
	\toprule
	& PoseCNN	&	BB8	& Tekin	& Heatmaps  & iPose & {\bf Ours} \\
	\midrule
	FPS &4&3&50&4&-&22\\
	\bottomrule
	\end{tabular}
}
	\vspace{-3mm}
	\caption{{\bf Runtime comparisons on Occluded-LINEMOD.} All methods run on a modern Nvidia GPU. \comment{, TitanX for BB8, Tekin and Ours, and GTX980Ti \MS{Is it comparable?} \YH{From Nvidia specs, GTX 980Ti may roughly have 62\% power of TitanX. So we only 3.1x times faster if taken this into account.} for Heatmaps.}}
	\label{tab:timings}
\end{table} 

% !TEX root = ../top.tex
% !TEX spellcheck = en-US

\begin{table}
	\centering
	\rowcolors{2}{gray!10}{white}
	\begin{tabular}{lrrrrr}
	\toprule
	&	NF & HC	& B-2 & B-10	&	Oracle	\\
	\midrule
	Ape			&37.8&58.2&58.2&59.1&{\bf 84.0} \\
	Can			&53.4&58.7&58.5&59.8&{\bf 89.0} \\
	Cat 		&42.6&46.1&47.4&46.9&{\bf 60.6} \\
	Driller		&52.5&56.8&59.4&59.0&{\bf 90.3} \\
	Duck		&40.4&42.8&42.4&42.6&{\bf 55.6} \\
	Eggbox$^*$	&{\bf 12.8}&11.2&12.1&11.9& 10.9 \\
	Glue$^*$	&14.7&15.8&15.1&16.5&{\bf 41.0} \\
	Holepun.	&58.4&62.2&63.1&63.6&{\bf 89.3} \\
	\midrule
	Average		&39.1&44.0&44.5&44.9&{\bf 65.1} \\
	\bottomrule
	FPS	& 26	& 26	&	25 & 22	& - \\
	\end{tabular}
	\vspace{-3mm}
	\caption{{\bf Accuracy (REP-5px) of different fusion strategies on Occluded-LINEMOD.} We compare a No-Fusion (NF) scheme with one that relies on the Highest-Confidence predictions, and with strategies relying on performing RANSAC on the $n$ most confident predictions (B-$n$). Oracle consists of choosing the best 2D location using the ground-truth one, and is reported to indicate the potential for improvement of our approach. In the bottom row, we also report the average runtime of these different strategies.}
	% represents the procedure choosing the closest 2D reprojection to the ground truth during testing. It shows that most of the ground truth positions have been covered by our candidates (see Sec.~\ref{sec:differentInference}). Average running time (in FPS) is also shown.}}
	\label{tab:different_inference}
\end{table}

\subsection{Inference Strategy}
\label{sec:inference}

At test time, given a query image, our network returns, for each foreground cell in the $S \times S$ grid of Section~\ref{eq:network}, an object class and a set of predicted 2D locations for the projection of the $N$ 3D keypoints. As we perform class-based segmentation instead of instance-based segmentation, there might  be ambiguities if two objects of the same class are present in the scene. To avoid that,  we leverage the fact that the predicted 2D keypoint locations tend to cluster according to the objects they correspond and use a simple pixel distance threshold to identify such clusters.

%More reliable multi-model fitting algorithms~\cite{Isack12,Magri16} should work better, but we do not investigate them further.

For each cluster, that is, for each object, we then exploit the confidence scores predicted by the network to establish 2D-to-3D correspondences between the image and the object's 3D model. The simplest way of doing so would be to use RANSAC on all the predictions. This, however, would significantly slow down our approach. Instead, we rely on the $n$ most confident 2D predictions for each 3D keypoint. In practice, we found $n=10$ to yield a good balance between speed and accuracy. Given these filtered 2D-to-3D correspondences, we obtain the 6D pose of each object using the RANSAC-based version of the EPnP algorithm of~\cite{Lepetit09}. Fig.~\ref{fig:multi_candidates} illustrates this procedure.

\section{Experiments}
We now evaluate our segmentation-driven multi-object 6D pose estimation method on the challenging Occluded-LINEMOD~\cite{Krull15} and YCB-Video~\cite{Xiang18b} datasets, which, unlike  LINEMOD~\cite{Hinterstoisser12b}, contain 6D pose annotations for each object appearing in all images.

% !TEX root = ../top.tex
% !TEX spellcheck = en-US

\begin{table*}
\centering
\begin{tabular}{lcccc!{\vrule width 1pt}cccccc}
	\toprule
	& \multicolumn{4}{c}{ADD-0.1d}	&	\multicolumn{6}{c}{REP-5px} \\
	&	Mask R-CNN & CPM &\cite{Xiang18b}& {\bf Ours} & 	Mask R-CNN  & CPM & \cite{Xiang18b}&\cite{Rad17}&\cite{Tekin18a}& {\bf Ours}	\\
	\midrule
	Average		& 11.8	& 12.7 &24.9&	{\bf 27.0}	& 22.4	& 22.9 &17.2&7.6&6.2	& {\bf 44.9} \\
	\bottomrule
\end{tabular}
\vspace{-3mm}
\caption{{\bf Comparison with human pose estimation methods on Occluded-LINEMOD.} We modified two state-of-the-art human pose estimation methods, Mask R-CNN~\cite{He17} and CPM~\cite{Wei16}, to output bounding box corner locations. 
%The global-inference 6D pose methods PoseCNN~\cite{Xiang18b}, BB8~\cite{Rad17}, and Tekin~\cite{Tekin18a} are also included. 
While both Mask R-CNN and CPM perform slightly better than other global-inference methods, our local approach yields much more accurate predictions.}
\label{tab:human_pose}
\end{table*}

\parag{Metrics.}

We report the commonly-used 2D reprojection (REP) error~\cite{Brachmann16a}. It encodes the average distance between the 2D reprojection of the 3D model points obtained using the predicted pose and those obtained with the ground-truth one. Furthermore, we also report the pose error in 3D space~\cite{Hinterstoisser12b}, which corresponds to the average distance between the 3D points transformed using the predicted pose and those obtained with the ground-truth one. As in~\cite{Li18a, Xiang18b}, we will refer to it as ADD. Since many objects in the datasets are symmetric,  we use the symmetric version of these two metrics and report their REP-5px and ADD-0.1d values. They assume the predicted pose to be correct if the REP is below a 5 pixel threshold and the ADD below 10\% of the model diameter, respectively. Below, we denote the objects that are considered to be symmetric by a $^*$ superscript. 

\parag{Implementation Details.}
As in~\cite{Redmon18}, we scale the input image to a 608 $\times$ 608 resolution for both training and testing. Furthermore, when regressing the 2D reprojections, we normalize the horizontal and vertical positions to the range $[0,10]$. We use the same normalization procedure when estimating the confidences.

We train the network for 300 epochs on Occluded-LINEMOD and 30 epochs on YCB-Video. In both cases, the initial learning rate is set to 1e-3, and is divided by 10 after 50\%, 75\%, and 90\% of the total number of epochs. We use SGD as our optimizer with a momentum of 0.9 and a weight decay of 5e-4. Each training batch contains 8 images, and we have employed the usual data augmentation techniques, such as random luminance, Gaussian noise, translation and scaling. We have also used the random erasing technique of~\cite{Zhong17} for better occlusion handling. Our source code is publicly available at \href{https://github.com/cvlab-epfl/segmentation-driven-pose}{https://github.com/cvlab-epfl/segmentation-driven-pose}.

% !TEX root = ../top.tex
% !TEX spellcheck = en-US

\begin{figure*}[htbp]
	\centering
	\begin{tabular}{ccc}
	\includegraphics[width=0.3 \linewidth, clip, trim=150 30 50 150]{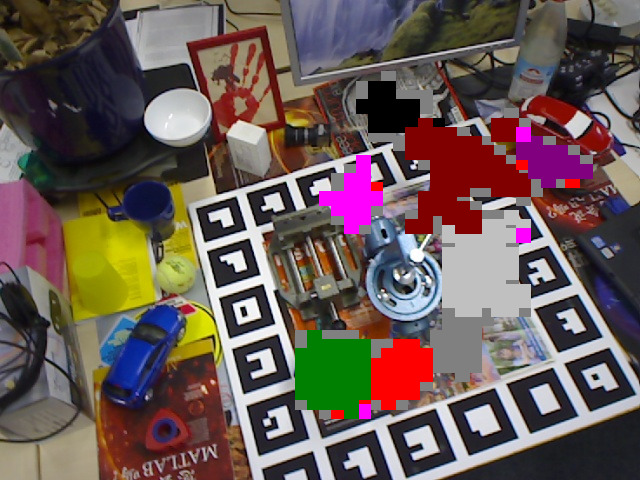} &
	\includegraphics[width=0.3 \linewidth, clip, trim=100 130 100 50]{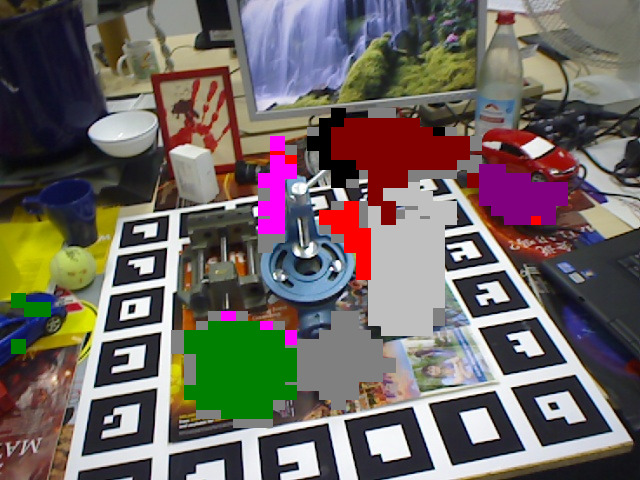} &
	\includegraphics[width=0.3 \linewidth, clip, trim=120 140 80 40]{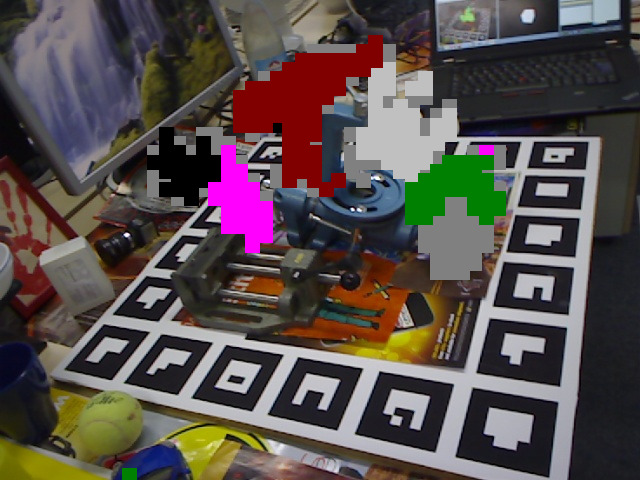} \\
	\includegraphics[width=0.3 \linewidth, clip, trim=150 30 50 150]{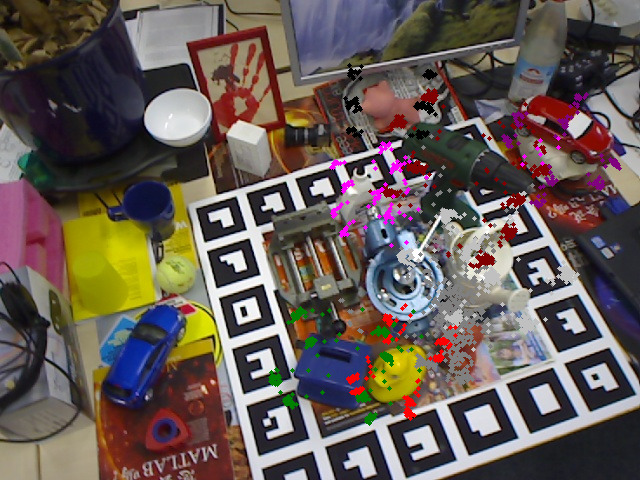} &
	\includegraphics[width=0.3 \linewidth, clip, trim=100 130 100 50]{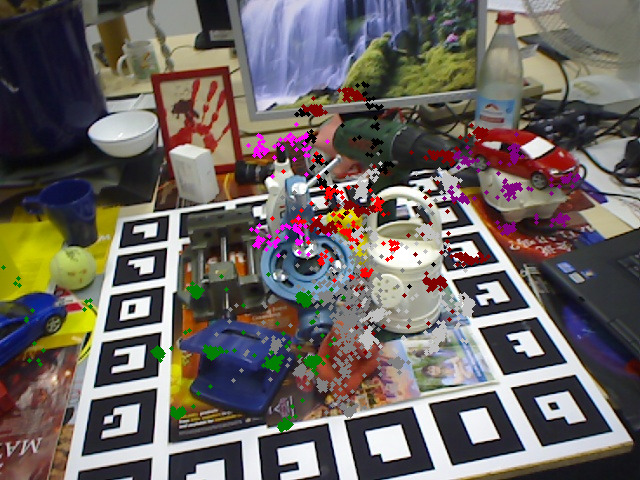} &
	\includegraphics[width=0.3 \linewidth, clip, trim=120 140 80 40]{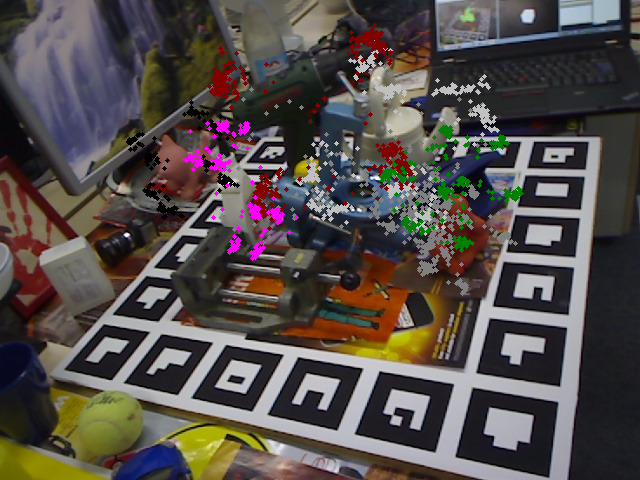} \\
	\includegraphics[width=0.3 \linewidth, clip, trim=150 30 50 150]{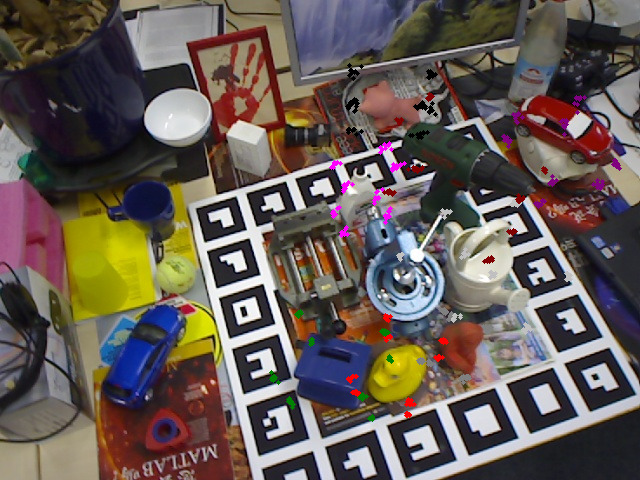} &
	\includegraphics[width=0.3 \linewidth, clip, trim=100 130 100 50]{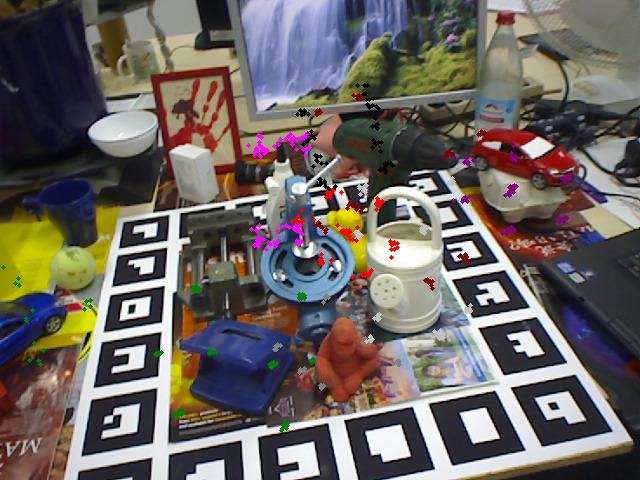} &
	\includegraphics[width=0.3 \linewidth, clip, trim=120 140 80 40]{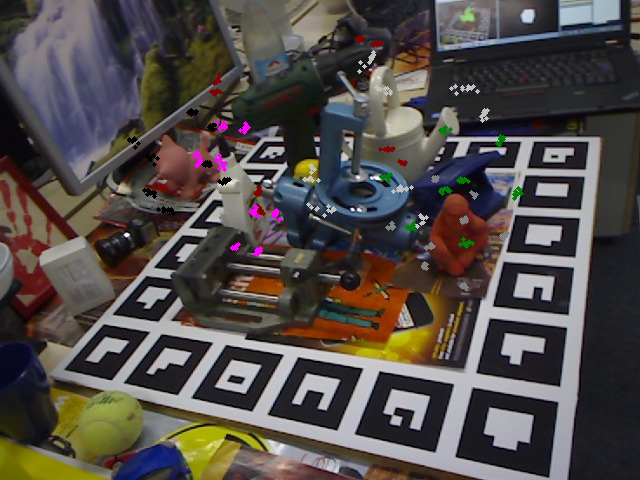} \\
	\includegraphics[width=0.3 \linewidth, clip, trim=150 30 50 150]{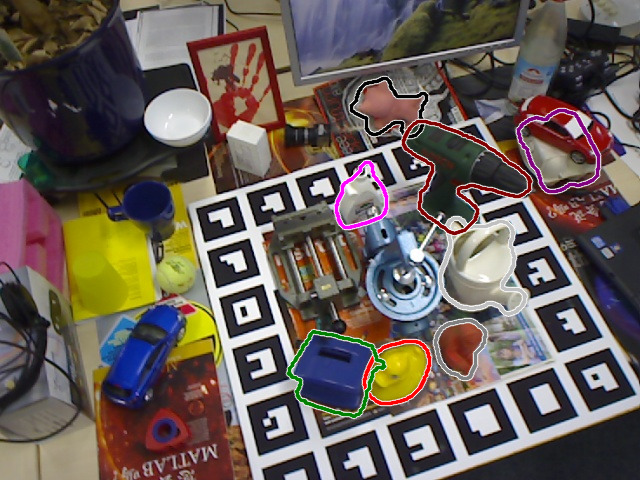} &
	\includegraphics[width=0.3 \linewidth, clip, trim=100 130 100 50]{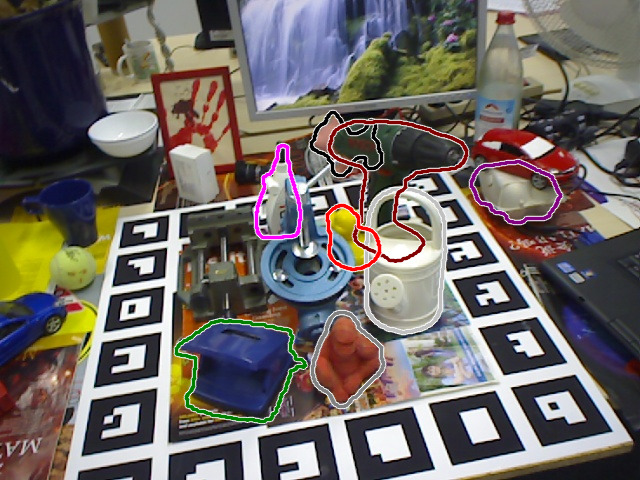} &
	\includegraphics[width=0.3 \linewidth, clip, trim=120 140 80 40]{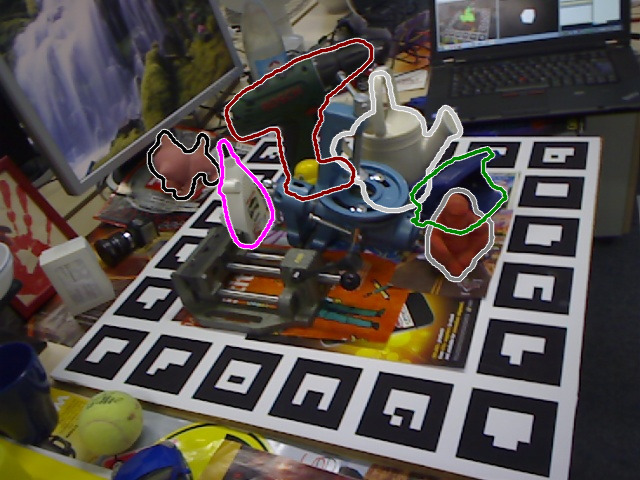} \\
	\end{tabular}
	\vspace{-2mm}
	\caption{{\bf Occluded-LINEMOD results}. In each column, we show, from top to bottom: the foreground segmentation mask, all 2D reprojection candidates, the selected 2D reprojections, and the final pose results. Our method generates accurate pose estimates, even in the presence of large occlusions. Furthermore, it can process multiple objects in real time.}
	\label{fig:results_visualization_occ}
\end{figure*}

\subsection{Evaluation on Occluded-LINEMOD}
\label{sec:eval_occlinemod}

The Occluded-LINEMOD dataset~\cite{Krull15} was compiled by annotating the pose of all the objects in a subset of the raw LINEMOD dataset~\cite{Hinterstoisser12b}. This subset depicts 8 different objects in 1214 images. Although depth information is also provided, we only exploit the RGB images. The Occluded-LINEMOD images, as the LINEMOD ones, depict a central object surrounded by non-central ones. The standard protocol consists of only evaluating on the {\it non-central} objects.

To create training data for our model, we follow the same procedure as in~\cite{Li18a,Tekin18a}. We use the mask inferred from the ground-truth pose to segment the {\it central} object in each image, since, as mentioned above, it will not be used for evaluation. We then generate synthetic images by inpainting between 3 and 8 objects on random PASCAL VOC images~\cite{Everingham10}. These objects are placed at random locations, orientations, and scales. This procedure still enables us to recover the occlusion state of each object and generate the corresponding segmentation mask. By using the central objects from any of the raw LINEMOD images, provided that it is one of the 8 objects used in Occluded-LINEMOD, we generated 20k training samples. 

%\YH{LINEMOD only annotate the central object for each image, even there are other types of objects around. It is rather simple because the central objects have no occlusions usually, and it does not support multi-object evaluation. Occluded-LINEMOD is an extended annotation of a subset of LINEMOD, which {\it only} annotates all the non-central objects for 1214 images (just a reverse way as in LINEMOD), and use all of them as the testing images. Training on LINEMOD and testing on Occluded-LINEMOD is a common way in 6D pose.}

\subsubsection{Comparing against the State of the Art}
We compare our method with the state-of-the-art ones of~\cite{Xiang18b} (PoseCNN),~\cite{Rad17} (BB8), and~\cite{Tekin18a} (Tekin), which all produce a single global pose estimate. Furthermore, we also report the results of the recent work of~\cite{Jafari18} (iPose), and~\cite{Oberweger18} (Heatmaps), which combines the predictions of multiple, relatively large patches,  but relies on an expensive sliding-window strategy. Note that~\cite{Oberweger18} also provides results obtained with the Feature Mapping technique~\cite{Rad18}. However, most methods, including ours, do not use this technique, and for a fair comparison, we therefore report the results of all methods, including that of~\cite{Oberweger18}, without it.

%\PF{You should probably say why not.} \YH{This Feature Mapping can transfer features of synthetic training images to the real test ones, which is helpful and is true performs better. Theoretically, it could improve the performance of most methods. But for a fair comparison, we do not use this data preparation technique.}

We report our results in Table~\ref{tab:occlinemod_eval} and provide the runtimes of the methods in Table~\ref{tab:timings}. Our method outperforms the global inference ones~\cite{Xiang18b,Rad17,Tekin18a} by a large margin. It also outperforms Heatmaps, albeit by a smaller one. Furthermore, thanks to our simple architecture and one-shot inference strategy, our method runs more than 5 times faster than Heatmaps. Our approach takes 30ms per-image for segmentation and 2D reprojection estimation, and 3-4ms per object for fusion. With 5 objects per image on average, this yields a runtime of about 50ms. Fig.~\ref{fig:results_visualization_occ} depicts some of our results. Note their accuracy even in the presence of large occlusions. 

%\YH{Table~\ref{tab:timings} also report their running times. In general, our method runs on a NVIDIA TITAN X. The segmentation and 2D reprojection estimation will consume 30ms per image, and the fusion needs about 3-4ms per object. Considering the fact that there are about 5 objects per image, our method only needs about 50ms for processing one image on average. }

\subsubsection{Comparison of Different Fusion Strategies}
\label{sec:differentInference}

As shown in Fig.~\ref{fig:multi_candidates}, not all local predictions of the 2D keypoint locations are accurate. Therefore, the fusion strategy based on the predicted confidence values that we described in Section~\ref{sec:inference} is important to select the right ones. Here, we evaluate its impact on the final pose estimate. To this end, we report the results obtained by taking the 2D location with highest confidence (HC) for each 3D keypoint and those obtained with different values $n$ in our $n$ most-confident selection strategy. We refer to this as B-$n$ for a particular value $n$. Note that we then use RANSAC on the selected 2D-to-3D correspondences.

In Table~\ref{tab:different_inference}, we compare the results of these different strategies with a fusion-free method that always uses the 2D reprojections predicted by the center grid, which we refer to as No-Fusion (NF). These results evidence that all fusion schemes outperform the No-Fusion one. We also report the Oracle results obtained by selecting the best predicted 2D location for each 3D keypoint using the ground truth 2D reprojections. This indicates that our approach could further benefit from improving the confidence predictions or designing a better fusion scheme. 

% !TEX root = ../top.tex
% !TEX spellcheck = en-US

\begin{table}
\centering
\scalebox{0.85}{
	\rowcolors{3}{gray!10}{white}
	\begin{tabular}{lrrr!{\vrule width 1pt}rrr}
		\toprule
		& \multicolumn{3}{c}{ADD-0.1d}	&	\multicolumn{3}{c}{REP-5px} \\
		& \cite{Xiang18b}	& \cite{Oberweger18} & {\bf Ours} & \cite{Xiang18b}	& \cite{Oberweger18} & {\bf Ours}	\\
		\midrule
{\small master\_chef\_can} & 3.6 & 32.9 & {\bf 33.0} & 0.1 & 9.9 & {\bf 21.0} \\
{\small cracker\_box} & 25.1 & {\bf 62.6} & 44.6 & 0.1 & {\bf 24.5} & 12.0 \\
{\small sugar\_box} & 40.3 & 44.5 & {\bf 75.6} & 7.1 & 47.0 & {\bf 56.3} \\
{\small tomato\_soup\_can} & 25.5 & 31.1 & {\bf 40.8} & 5.2 & 41.5 & {\bf 46.2} \\
{\small mustard\_bottle} & 61.9 & 42.0 & {\bf 70.6} & 6.4 & 42.3 & {\bf 70.3} \\
{\small tuna\_fish\_can} & 11.4 & 6.8 & {\bf 18.1} & 3.0 & 7.1 & {\bf 39.3} \\
{\small pudding\_box} & 14.5 & {\bf 58.4} & 12.2 & 5.1 & {\bf 43.9} & 17.3 \\
{\small gelatin\_box} & 12.1 & 42.5 & {\bf 59.4} & 15.8 & 62.1 & {\bf 83.6} \\
{\small potted\_meat\_can} & 18.9 & {\bf 37.6} & 33.3 & 23.1 & 38.5 & {\bf 60.7} \\
{\small banana} & {\bf 30.3} & 16.8 & 16.6 & 0.3 & 8.2 & {\bf 22.4} \\
{\small pitcher\_base} & 15.6 & 57.2 & {\bf 90.0} & 0 & 15.9 & {\bf 33.5} \\
{\small bleach\_cleanser} & 21.2 & 65.3 &{\bf  70.9} & 1.2 & 12.1 & {\bf 43.3} \\
{\small bowl}$^*$ & 12.1 & 25.6 & {\bf 30.5} & 4.4 & {\bf 16.0} & 13.3 \\
{\small mug} & 5.2 & 11.6 & {\bf 40.7} & 0.8 & 20.3 & {\bf 38.1} \\
{\small power\_drill} & 29.9 & 46.1 &{\bf  63.5} & 3.3 & 40.9 & {\bf 43.3} \\
{\small wood\_block}$^*$ & 10.7 & {\bf 34.3} & 27.7 & 0 & {\bf 2.5} & {\bf 2.5} \\
{\small scissors} & 2.2 & 0 & {\bf 17.1} & 0 & 0 & {\bf 8.8} \\
{\small large\_marker} & 3.4 & 3.2 & {\bf 4.8} & 1.4 & 0 & {\bf 13.6} \\
{\small large\_clamp}$^*$ & {\bf 28.5} & 10.8 & 25.6 & 0.3 & 0 & {\bf 7.6} \\
{\small extra\_large\_clamp}$^*$ & 19.6 & {\bf 29.6} & 8.8 & {\bf 0.6} & 0 & {\bf 0.6} \\
{\small foam\_brick}$^*$ & {\bf 54.5} & 51.7 & 34.7 & 0 & {\bf 52.4} & 13.5 \\
		\midrule
		Average		& 21.3	&	33.6	&	{\bf 39.0}	& 3.7	& 23.1	& {\bf 30.8} \\
		\bottomrule
	\end{tabular}
}
	\vspace{-3mm}
	\caption{{\bf Comparison with the state of the art on YCB-Video.} We compare our results with those of PoseCNN~\cite{Xiang18b} and Heatmaps~\cite{Oberweger18}.}
	\label{tab:ycbvideo_eval}
\end{table}

% !TEX root = ../top.tex
% !TEX spellcheck = en-US

\begin{figure*}[htbp]
	\centering
	\begin{tabular}{ccc}
	\includegraphics[width=0.3 \linewidth, clip, trim=30 0 70 120]{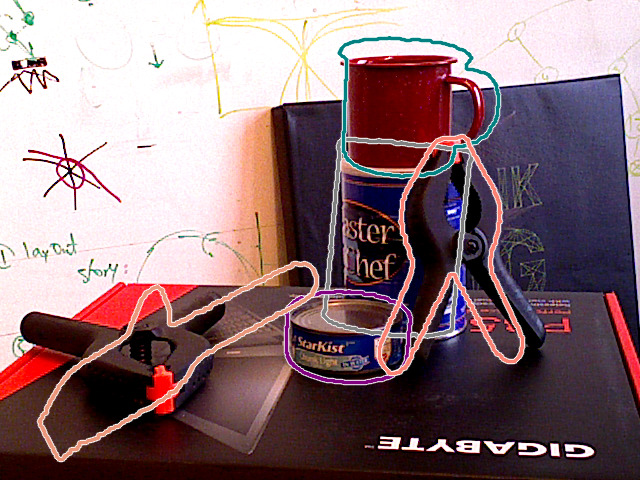}&
	\includegraphics[width=0.3 \linewidth, clip, trim=0 0 100 120]{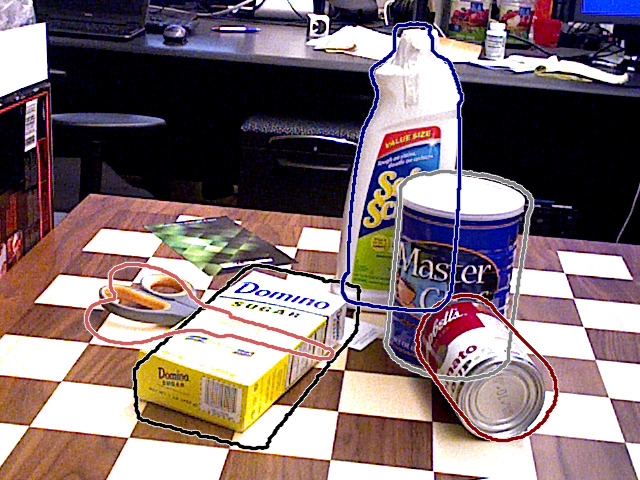}&
	\includegraphics[width=0.3 \linewidth, clip, trim=40 0 60 120]{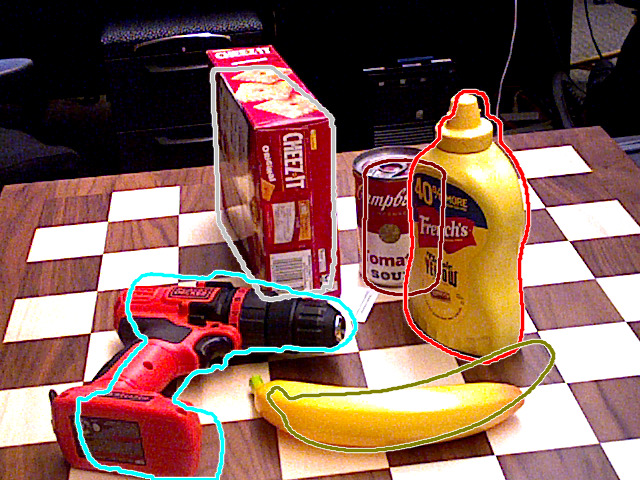}\\
	\includegraphics[width=0.3 \linewidth, clip, trim=30 0 70 120]{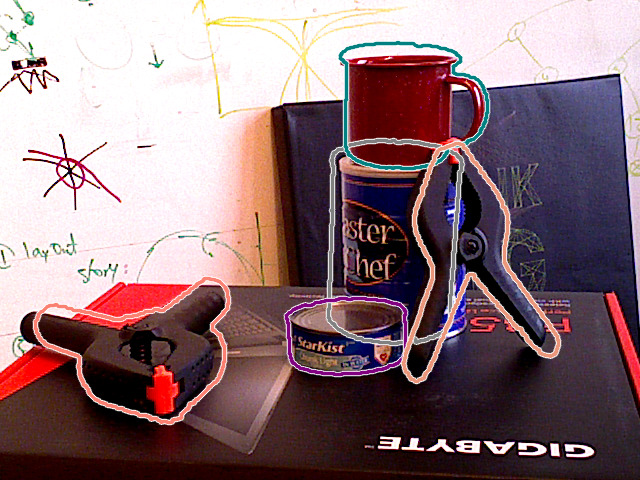}&
	\includegraphics[width=0.3 \linewidth, clip, trim=0 0 100 120]{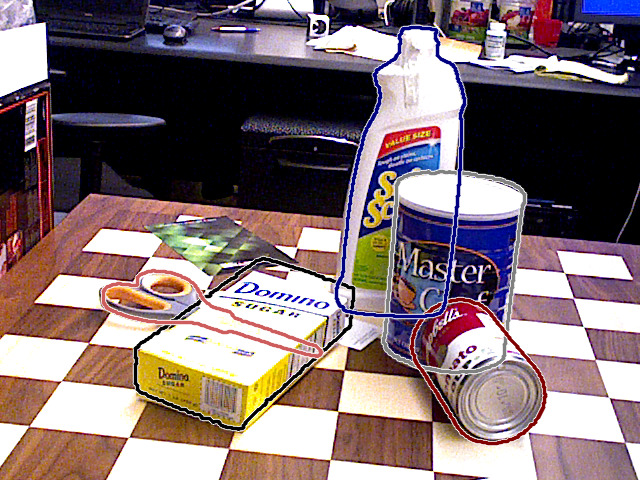}&
	\includegraphics[width=0.3 \linewidth, clip, trim=40 0 60 120]{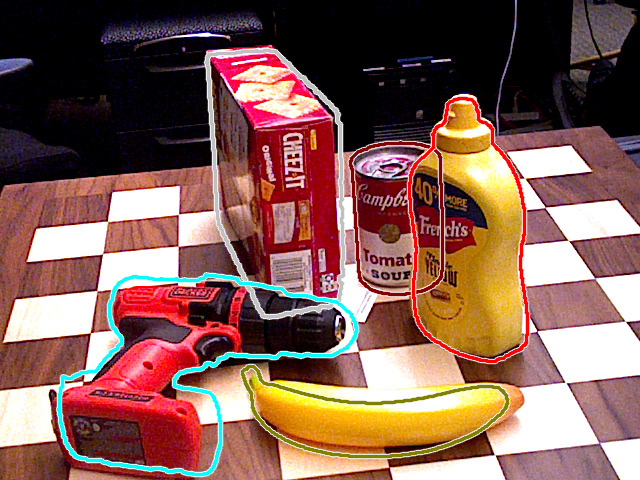}
	\end{tabular}
	\vspace{-2mm}
	\caption{{\bf Comparison to PoseCNN~\cite{Xiang18b} on YCB-Video.} (Top) PoseCNN and (Bottom) Our method. This demonstrates the benefits of reasoning about local object parts instead of globally, particularly in the presence of large occlusions.}
	\label{fig:results_visualization_ycb}
\end{figure*}

\subsubsection{Comparison with Human Pose Methods}
Our method enables us to infer keypoints' locations of rigid objects from local visible object regions and does not require the more global information extracted from larger receptive fields that are more sensitive to occlusions. To further back up this claim, we compare our approach to two state-of-the-art human pose estimation methods, Mask R-CNN~\cite{He17} and Convolutional Pose Machines (CPM)~\cite{Wei16}, which target non-rigid objects, {\it i.e.} human bodies. By contrast, dealing with rigid objects allows us to rely on local predictions that can be robustly combined. Specifically, we modified the publicly available code of Mask R-CNN and CPM to output 8 bounding box 2D corners instead of human keypoints and trained these methods on Occluded-LINEMOD. As shown in Table~\ref{tab:human_pose}, while both Mask R-CNN and CPM perform slightly better than other \emph{global-inference} methods, our \emph{local} approach yields much more accurate predictions.

\subsection{Evaluation on YCB-Video}

We also evaluate our method on the recent and more challenging YCB-Video dataset~\cite{Xiang18b}. It comprises 21 objects taken from the YCB dataset~\cite{Calli15,Calli17}, which are of diverse sizes and with different degrees of texture. This dataset contains about 130K real images from 92 video sequences, with an additional 80K synthetically rendered images that only contain foreground objects. It provides the pose annotations of all the objects, as well as the corresponding segmentation masks. The test images depict a great diversity in illumination, noise, and occlusions, which makes this dataset extremely challenging. As before, while depth information is available, we only use the color images. Here, we generate complete synthetic images from the 80K synthetic foreground ones by using the same random background procedure as in Section~\ref{sec:eval_occlinemod}. As before, we report results without feature mapping, because neither PoseCNN nor our approach use them. %\PF{Again, why not?}

\subsubsection{Comparing against the State of the Art}

Fewer methods have reported results on this newer dataset. In Table~\ref{tab:ycbvideo_eval}, we contrast our method with the two baselines that have. Our method clearly outperforms both PoseCNN~\cite{Xiang18b} and Heatmaps~\cite{Oberweger18}. Furthermore, recall that our approach runs more than 5 times faster than either of them.

In Fig.~\ref{fig:results_visualization_ycb}, we compare qualitative results of PoseCNN and ours. While our pose estimates are not as accurate on this dataset as on Occluded-LINEMOD, they are still much better than those of PoseCNN. Again, this demonstrates the benefits of reasoning about local object parts instead of globally, particularly in the presence of large occlusions.

\subsection{Discussion}
\label{sec:limitation}

Although our method performs well in most cases, it still can handle neither the most extreme occlusions nor tiny objects. In such cases, the grid we rely on becomes to rough a representation. This, however, could be addressed by using a finer grid, or, to limit the computational burden, a grid that is adaptively subdivided to better handle each image region. Furthermore, as shown in Table~\ref{tab:different_inference}, we do not yet match the performance of an oracle that chooses the best predicted 2D location for each 3D keypoint. This suggests that there is room to improve the quality of the predicted confidence score, as well as the fusion procedure itself. This will be the topic of our future research.

% !TEX root = ../top.tex
% !TEX spellcheck = en-US

\section{Conclusion}
\label{sec:conclusion}

We have introduced a segmentation-driven approach to 6D object pose estimation, which jointly detects multiple objects and estimates their pose. By combining multiple local pose estimates in a robust fashion, our approach produces accurate results without the need for a refinement step, even in the presence of large occlusions. Our experiments on two challenging datasets have shown that our approach outperforms the state of the art, and, as opposed to the best competitors, predicts the pose of multiple objects in real time. In the future, we will investigate the use of other backbone architectures for the encoder and devise a better fusion strategy to select the best predictions before performing PnP. We will also seek to incorporate the PnP step of our approach into the network, so as to have a complete, end-to-end learning framework.

\vspace{1em}
{\bf Acknowledgments} This work was supported in part by the Swiss Innovation Agency Innosuisse. We would like to thank Markus Oberweger and Yi Li for clarifying details about their papers, and Zheng Dang for helpful discussions.

{\small
\bibliographystyle{ieee}
\bibliography{string,vision,learning}
}

\end{document}